\documentclass[10pt,logo,twocolumn,copyright]{nvidia}

\usepackage{color,xcolor}
\usepackage{epsfig}
\usepackage{graphicx}

\usepackage{adjustbox}
\usepackage{array}
\usepackage{booktabs}
\usepackage{colortbl}
\usepackage{wrapfig}
\usepackage{hhline}
\usepackage{multirow}
\usepackage{subcaption}
\usepackage[toc,page]{appendix}

\usepackage{amsmath,amsfonts,amsthm,amssymb}
\usepackage{bm}
\usepackage{nicefrac}
\usepackage{microtype}
\usepackage{inconsolata}
\usepackage{pifont}
\usepackage{changepage}
\usepackage{extramarks}
\usepackage{fancyhdr}
\usepackage{lastpage}
\usepackage{setspace}
\usepackage{soul}
\usepackage{xspace}
\usepackage{indentfirst}
\usepackage{paralist}
\usepackage{arydshln}
\usepackage{makecell}

\usepackage[breaklinks=true,colorlinks,allcolors=blue,bookmarks=false]{hyperref}
\usepackage{url}

\usepackage{algorithm,algorithmic}
\usepackage{enumerate}
\usepackage{lipsum}
\usepackage{tikz}
\usetikzlibrary{tikzmark}

\newcolumntype{L}[1]{>{\raggedright\let\newline\\\arraybackslash\hspace{0pt}}m{#1}}
\newcolumntype{C}[1]{>{\centering\let\newline\\\arraybackslash\hspace{0pt}}m{#1}}
\newcolumntype{R}[1]{>{\raggedleft\let\newline\\\arraybackslash\hspace{0pt}}m{#1}} 
\newcommand{\xpar}[1]{\noindent\textbf{#1}\ \ }

\newcommand{\sect}[1]{Section~\ref{#1}}

\newcommand{\fig}[1]{Figure~\ref{#1}}

\newcommand{\tab}[1]{Table~\ref{#1}}

\newcommand{\ignorethis}[1]{}

\makeatletter
\DeclareRobustCommand\onedot{\futurelet\@let@token\@onedot}
\def\@onedot{\ifx\@let@token.\else.\null\fi\xspace}

\def\eg{\emph{e.g}\onedot}

\makeatother

\makeatletter
\def\adl@drawiv#1#2#3{%
        \hskip.5\tabcolsep
        \xleaders#3{#2.5\@tempdimb #1{1}#2.5\@tempdimb}%
                #2\z@ plus1fil minus1fil\relax
        \hskip.5\tabcolsep}
\newcommand{\cdashlinelr}[1]{%
  \noalign{\vskip\aboverulesep
           \global\let\@dashdrawstore\adl@draw
           \global\let\adl@draw\adl@drawiv}
  \cdashline{#1}
  \noalign{\global\let\adl@draw\@dashdrawstore
           \vskip\belowrulesep}}
\makeatother

\newcommand{\cmark}{\ding{51}}
\newcommand{\xmark}{\ding{55}}

\definecolor{nvcolor}{HTML}{76b900}
\definecolor{mydarkblue}{rgb}{0,0.08,1}
\definecolor{mydarkgreen}{rgb}{0.02,0.6,0.02}
\definecolor{mydarkred}{rgb}{0.8,0.02,0.02}
\definecolor{mydarkorange}{rgb}{0.40,0.2,0.02}
\definecolor{mypurple}{RGB}{111,0,255}
\definecolor{myred}{rgb}{1.0,0.0,0.0}
\definecolor{mygold}{rgb}{0.75,0.6,0.12}
\definecolor{mydarkgray}{rgb}{0.66, 0.66, 0.66}

\definecolor{darkblue}{rgb}{0,0.08,1}
\definecolor{darkgreen}{rgb}{0.02,0.6,0.02}
\definecolor{darkred}{rgb}{0.8,0.02,0.02}
\definecolor{darkorange}{rgb}{0.40,0.2,0.02}
\definecolor{darkpurple}{RGB}{111,0,255}

\newif\ifdraft
\draftfalse 

\newcommand{\stwo}{{S\textsuperscript{2}}\xspace}
\newcommand{\dynstwo}{{Dynamic-S\textsuperscript{2}}\xspace}

\begin{document}

\title{NVILA: Efficient Frontier Visual Language Models}

\author{Zhijian Liu\textsuperscript{1,$\dag$} \qquad Ligeng Zhu\textsuperscript{1,$\dag$} \qquad Baifeng Shi\textsuperscript{3} \qquad Zhuoyang Zhang\textsuperscript{2} \qquad Yuming Lou\textsuperscript{6} \qquad Shang Yang\textsuperscript{2} \qquad Haocheng Xi\textsuperscript{3} \qquad Shiyi Cao\textsuperscript{3} \qquad Yuxian Gu\textsuperscript{2,6} \qquad Dacheng Li\textsuperscript{3} \qquad\qquad Xiuyu Li\textsuperscript{3} \qquad Yunhao Fang\textsuperscript{4} \qquad Yukang Chen\textsuperscript{1} \qquad Cheng-Yu Hsieh\textsuperscript{5} \qquad De-An Huang\textsuperscript{1} \qquad An-Chieh Cheng\textsuperscript{4} \qquad 
Vishwesh Nath\textsuperscript{1} \qquad Andriy Myronenko\textsuperscript{1} \qquad Jinyi Hu\textsuperscript{2,6} \qquad Sifei Liu\textsuperscript{1} \qquad Ranjay Krishna\textsuperscript{5} \qquad 
Daguang Xu\textsuperscript{1} \qquad 
Xiaolong Wang\textsuperscript{1,4}  \qquad Pavlo Molchanov\textsuperscript{1} \qquad\qquad\qquad Jan Kautz\textsuperscript{1} \qquad Hongxu Yin\textsuperscript{1,$\ddag$} \qquad Song Han\textsuperscript{1,2,$\ddag$} \qquad Yao Lu\textsuperscript{1,$\dag$,$\ddag$} \\~\\
\textsuperscript{1}NVIDIA \quad \textsuperscript{2}MIT \quad \textsuperscript{3}UC Berkeley \quad \textsuperscript{4}UC San Diego \quad \textsuperscript{5}University of Washington \quad \textsuperscript{6}Tsinghua University \\
\textsuperscript{$\dag$}Equal contribution \quad \textsuperscript{$\ddag$}Equal advisory}

\begin{abstract}

\textbf{Abstract:} \hspace{2pt}
Visual language models (VLMs) have made significant advances in accuracy in recent years. However, their efficiency has received much less attention. This paper introduces \textbf{NVILA}, a family of open VLMs designed to jointly optimize efficiency and accuracy. Building on top of VILA, we improve its model architecture by first \textbf{scaling up} the spatial and temporal resolutions, and then \textbf{compressing} visual tokens. This ``\textit{scale-then-compress}'' approach enables NVILA to efficiently process high-resolution images and long videos. We further conduct a systematic investigation that enhances NVILA's efficiency throughout its entire lifecycle, from training and fine-tuning to deployment. NVILA matches or surpasses the accuracy of leading open and proprietary VLMs across a wide range of image and video benchmarks. At the same time, it reduces training cost by \textbf{1.9--5.1$\times$}, prefilling latency by \textbf{1.6--2.2$\times$}, and decoding latency by \textbf{1.2--2.8$\times$}. We release our code and models to facilitate reproducibility.

\vspace{5pt}

\textbf{Links:} \hspace{2pt} \href{https://github.com/NVlabs/VILA}{Code} (on GitHub) | \href{https://huggingface.co/collections/Efficient-Large-Model/nvila-674f8163543890b35a91b428}{Models} (on Hugging Face)

\vspace{10pt}

\begin{center}
    \includegraphics[width=\linewidth]{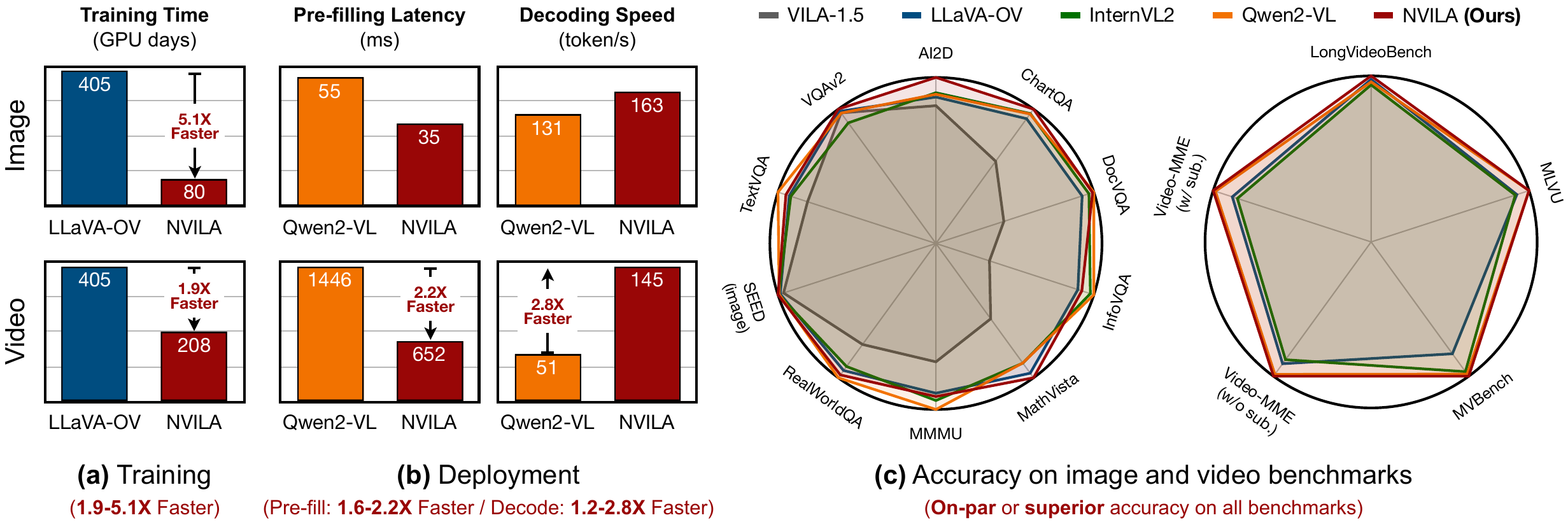}
    \captionof{figure}{\textbf{NVILA -- Efficient Frontier VLMs}. \textbf{(a)} NVILA trains image and video models 5.1$\times$ and 1.9$\times$ faster, respectively, than LLaVA-OneVision (OV), the only baseline with publicly reported training costs. \textbf{(b)} Against Qwen2-VL, NVILA achieves a 1.6--2.2$\times$ speedup in the prefilling stage and a 1.2--2.8$\times$ speedup in the decoding stage. \textbf{(c)} This efficiency comes at no cost to accuracy; in fact, NVILA delivers comparable or superior accuracy across image and video benchmarks. All models in this figure have 8B parameters. Training time in (a) is measured on NVIDIA H100 GPUs; inference speed in (b) is measured on a single NVIDIA RTX 4090 GPU. Accuracy in (c) is normalized to the highest score on each benchmark.}
    \label{fig:teaser}
\end{center}

\end{abstract}

\maketitle

\begin{figure*}[!t]
    \centering
    \includegraphics[width=0.9\linewidth]{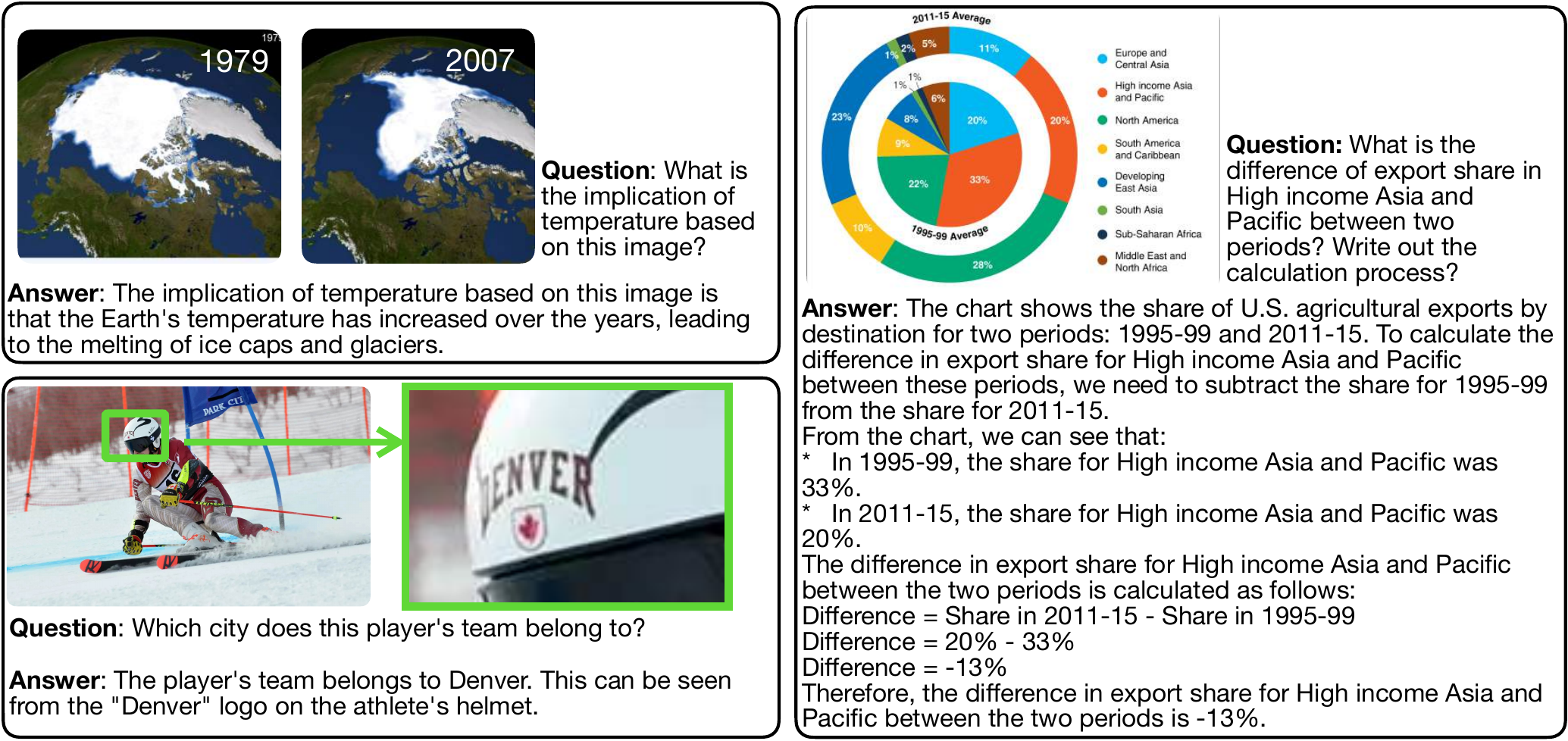}
    \includegraphics[width=0.9\linewidth]{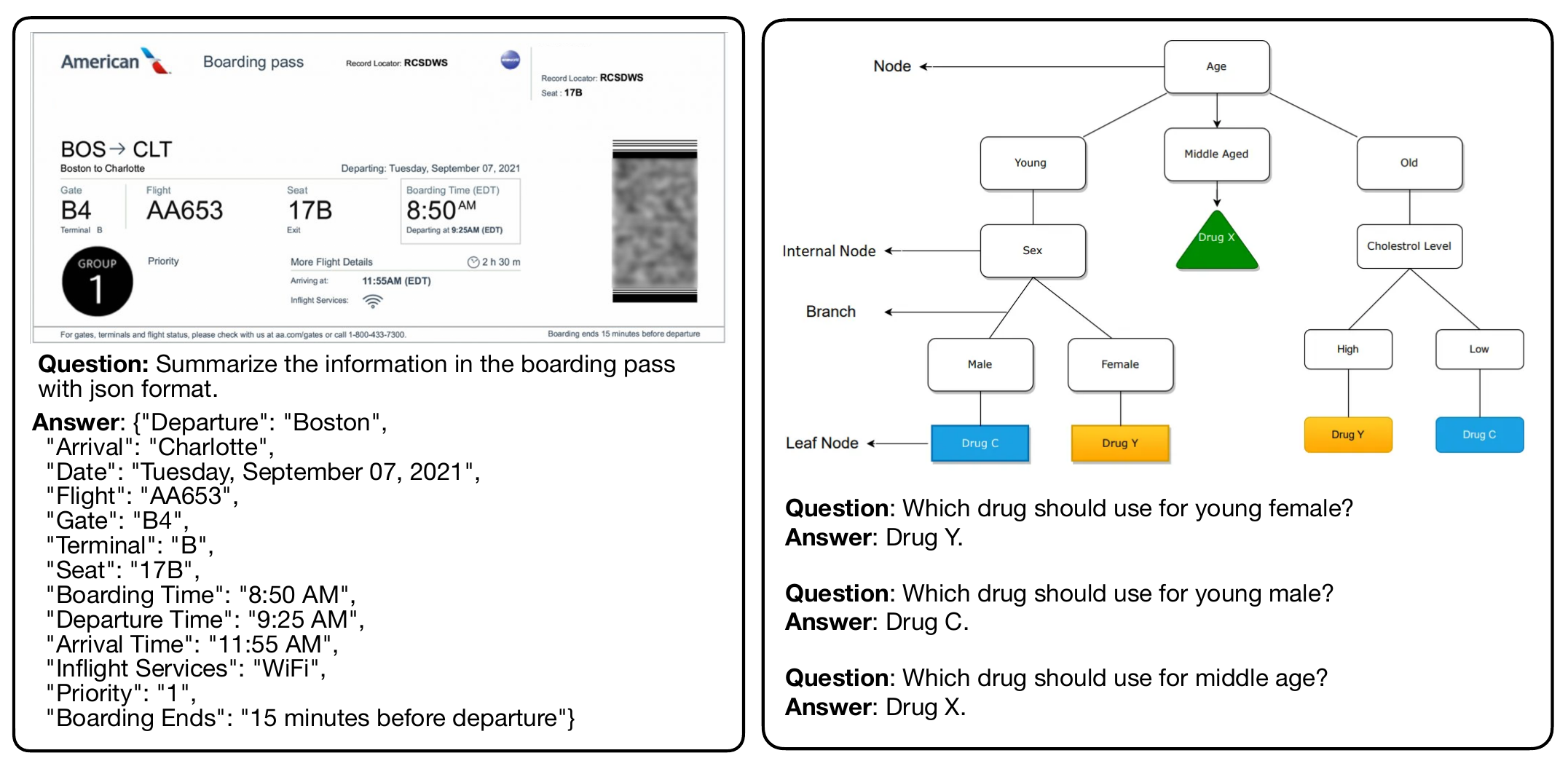}
    \includegraphics[width=0.9\linewidth]{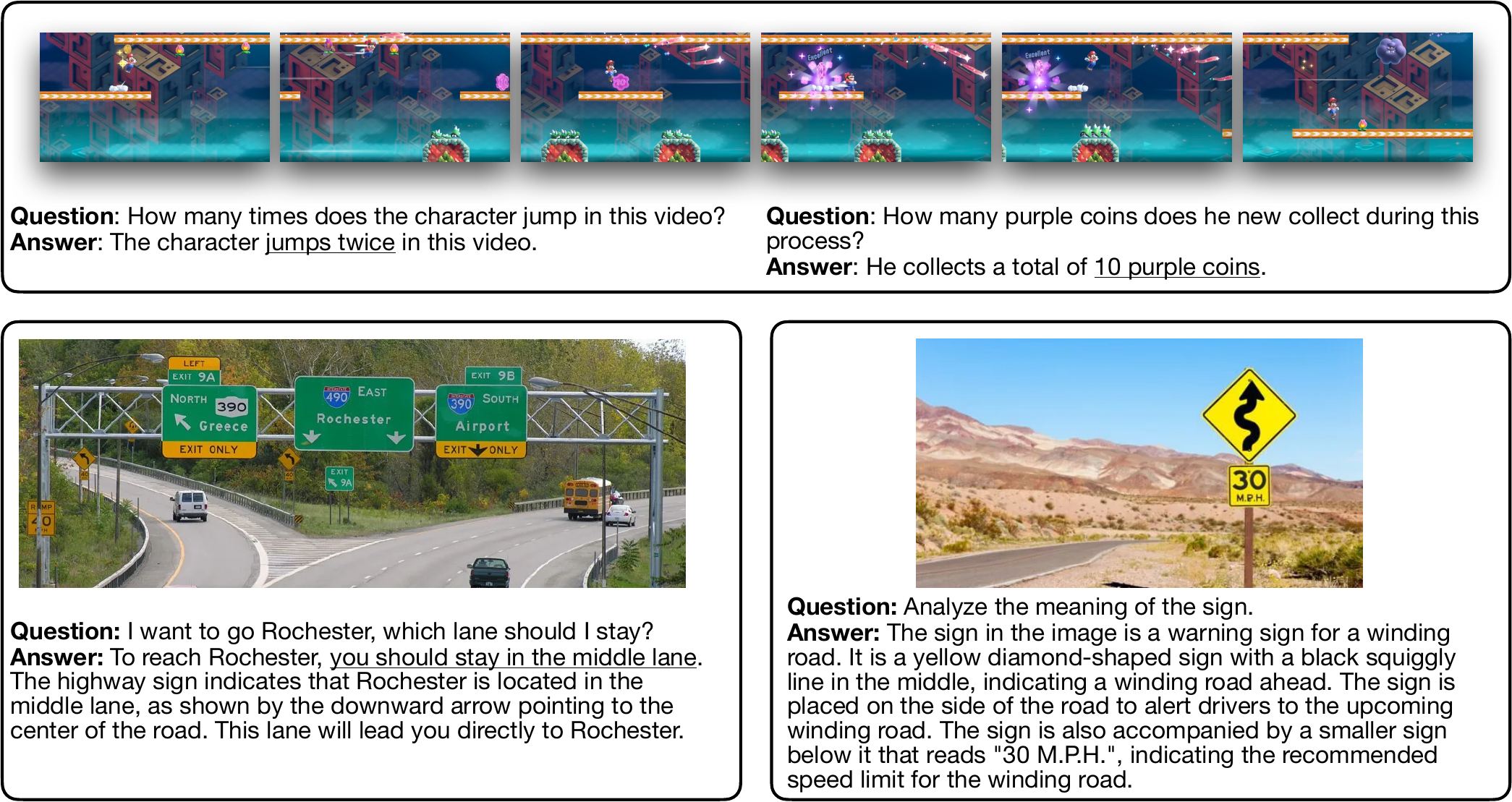}
    \caption{\textbf{Qualitative examples}. NVILA produces accurate responses across a diverse set of tasks, including OCR on text-heavy documents, multi-step visual reasoning, and multi-image comparison.}
    \label{fig:qualitative_examples}
\end{figure*}

\begin{figure}[t]
    \centering
    \includegraphics[width=\linewidth]{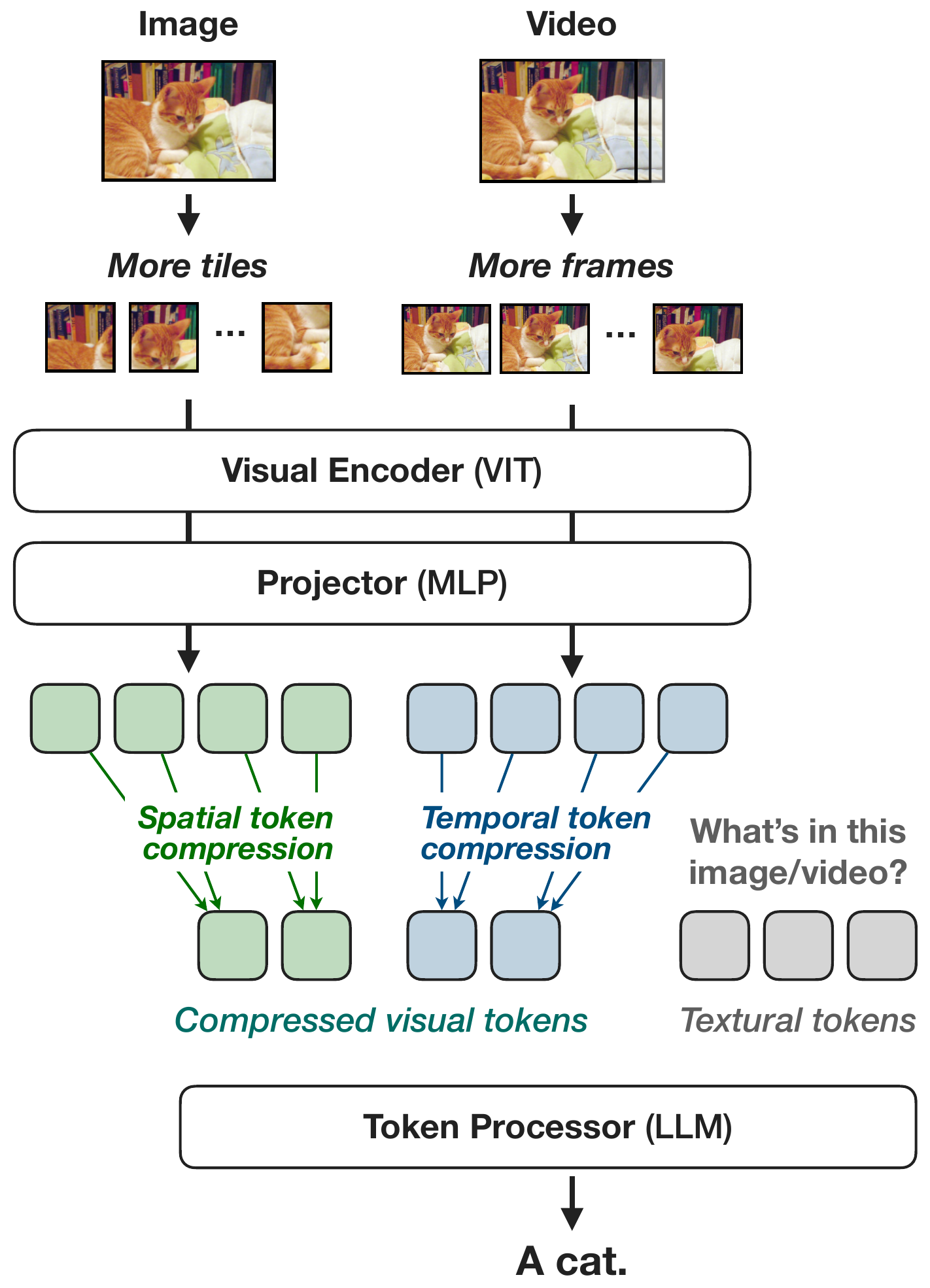}
    \caption{\textbf{NVILA architecture}. NVILA consists of three components: a visual encoder (SigLIP), a projector (two-layer MLP), and a token processor (Qwen2 LLM). Input images are processed by \dynstwo tiling and spatially compressed before being fed to the LLM alongside text tokens.}
    \label{fig:overview}
\end{figure}

\section{Introduction}

Visual language models (VLMs) integrate visual and textual inputs, enabling a wide range of vision-language interactions. In recent years, the research community has made substantial progress in improving their accuracy~\cite{liu2024llava,lin2024vila,chen2024internvl,li2024llavaonevision,wang2024qwen2vl} and broadening their applications across diverse domains, including robotics~\cite{brohan2022rt1,zhang2024navid,cheng2024navila}, autonomous driving~\cite{tian2024drivevlm}, and medical applications~\cite{saab2024medgemini,nath2024vilam3}. However, much less attention has been paid to their efficiency.

VLMs are expensive across multiple dimensions. First, \textit{training a VLM is time-consuming}: training a state-of-the-art 7B VLM~\cite{li2024llavaonevision} can take up to 400 GPU days, with even higher costs for larger models. This creates a significant entry barrier for researchers. Second, VLMs often require adaptation to specialized domains (\eg, medical imaging), but \textit{fine-tuning a VLM is memory-intensive}: fully fine-tuning a 7B VLM can require over 64\,GB of GPU memory, well beyond what is available on most consumer-grade GPUs. Finally, VLMs are increasingly deployed in edge applications with a tight compute budget (\eg, laptops, robots), so \textit{deploying a VLM is latency-sensitive}. Addressing all three together requires a comprehensive approach to VLM efficiency.

In this paper, we introduce \textbf{NVILA}, a family of open VLMs designed to jointly optimize efficiency and accuracy. Building on VILA~\cite{lin2024vila}, we improve its model architecture by first scaling up the spatial and temporal resolutions, and then compressing visual tokens. \textit{Scaling} preserves more visual detail, raising the accuracy ceiling, while \textit{compression} reduces the visual token count, lowering computational cost. This ``\textit{scale-then-compress}'' strategy enables NVILA to process high-resolution images and long videos both accurately and efficiently. We further systematically optimize efficiency across NVILA's full lifecycle: training, fine-tuning, and deployment.

These innovations make NVILA both efficient and accurate. It reduces training cost by \textbf{1.9--5.1$\times$}, prefilling latency by \textbf{1.6--2.2$\times$}, and decoding latency by \textbf{1.2--2.8$\times$}. It also matches or surpasses the accuracy of leading open VLMs~\cite{wang2024qwen2vl,chen2024internvl,lin2024vila} and proprietary VLMs~\cite{openai2024gpt4o,anthropic2024claude35} across a wide range of image and video benchmarks. NVILA further opens up new application domains, including temporal localization, robotic navigation, and medical imaging. We release our code and models to support reproducibility and to inspire further work on efficient VLMs.

\section{Approach}

We first describe NVILA’s efficient model architecture, which follows a ``\textit{scale-then-compress}'' design: \textit{scaling up} spatial and temporal resolutions to raise the accuracy ceiling, then \textit{compressing} visual tokens to recover efficiency. We then present strategies to improve efficiency across NVILA’s full lifecycle: training, fine-tuning, and deployment. Unless otherwise specified, all analysis uses the 8B model.

\subsection{Efficient Model Architecture}
\label{sec:approach:efficient_arch}

We build NVILA on top of VILA~\cite{lin2024vila}, an autoregressive VLM. As shown in \fig{fig:overview}, it consists of three components: a \textit{visual encoder} that extracts features from visual inputs (\eg, images, videos); a \textit{projector} that aligns embeddings across visual and language modalities; and a \textit{token processor}, typically instantiated with an LLM, which takes both visual and language tokens as input and produces language tokens. Specifically, NVILA uses SigLIP~\cite{zhai2023siglip} as its visual encoder, a two-layer MLP as its projector, and Qwen2~\cite{yang2024qwen2} (in various parameter sizes) as its token processor.

VILA processes images at a fixed 448$\times$448 resolution regardless of aspect ratio, and samples at most 14 frames per video\footnote{This is the configuration for VILA-1.5 40B; other variants, such as VILA-1.5 3B, use 384$\times$384 resolution and 8 frames.}. Both choices introduce significant information loss and limit the model's ability to handle high-resolution images and long videos. As a result, VILA lags behind leading VLMs, especially on text-heavy image and long-video benchmarks (\tab{tab:results:image} and \tab{tab:results:video}).

We therefore advocate for the ``\textit{scale-then-compress}'' paradigm: first \textit{scale up} the spatial and temporal resolutions to raise the \textit{accuracy} ceiling, then \textit{compress} the visual tokens to recover \textit{efficiency}. Scaling alone, however, significantly increases compute: doubling the resolution quadruples the number of visual tokens, and self-attention's quadratic complexity in the LLM further amplifies the cost. Compression then offsets this overhead, since higher information density lets the model retain (or even surpass) the detail captured at lower resolution while using fewer total tokens.

\begin{table*}[t]
    \renewcommand{\arraystretch}{1.1}
    \footnotesize\centering
    \caption{\textbf{Spatial ``scale-then-compress''}. Scaling the spatial resolution with \dynstwo greatly improves accuracy, particularly on text-heavy benchmarks. Compressing visual tokens via spatial pooling reduces both tile count and tokens-per-tile at a moderate accuracy cost. Adding a visual encoder pre-training (VEP) stage recovers most of this loss. \textit{IM-10} denotes the average score across the 10 image benchmarks in \tab{tab:results:image}.}
    \begin{tabular}{lclccccc}
        \toprule
        & \makecell{Spatial\\Pooling} & \#Tokens/Tile & \#Tiles/Image & AI2D & DocVQA & TextVQA & IM-10 \\
        \midrule
        Baseline (VILA-1.5) & 2$\times$2 & 256 (=16$\times$16) & 1 & 87.0 & 61.3 & 67.5 & 61.2 \\
        \midrule
        \textbf{Scale} (\dynstwo) & 2$\times$2 & 256 (=16$\times$16) & 9--12 & 90.1 & 91.1 & 77.0 & 71.5 \\
        \textbf{Scale} + \textbf{Compress} & 3$\times$3 & 121 (=11$\times$11) & 1--12 & 87.4 & 82.3 & 74.1 & 67.1 \\
        \textbf{Scale} + \textbf{Compress} + VEP & 3$\times$3 & 121 (=11$\times$11) & 1--12 & 89.8 & 88.8 & 76.1 & 70.8 \\
        \midrule
        \rowcolor{black!10} \multicolumn{8}{l}{\textit{Alternative Designs}} \\
        TokenLearner      & -- & 121 & 1--12 & 90.0 & 86.5 & 75.6 & 69.8 \\
        Perceiver Resampler      & -- & 121 & 1--12 & 76.8 & 71.8 & 65.3 & 59.4 \\
        \bottomrule
    \end{tabular}
    \label{tab:compression:spatial}
\end{table*}
\begin{table*}[t]
    \renewcommand{\arraystretch}{1.1}
    \footnotesize\centering
    \caption{\textbf{Temporal ``scale-then-compress''}. \textit{Scaling up} the number of frames consistently improves video understanding. \textit{Compressing} visual tokens via temporal averaging reduces token count substantially with only a marginal accuracy drop, allowing NVILA to process up to 256 frames at a similar token budget.}
    \label{tab:compression:temporal}
    \begin{tabular}{lcclcccc}
        \toprule
        & \multirow{2.5}{*}{\#Frames} & \multirow{2.5}{*}{\makecell{Temporal\\Pooling}} & \multirow{2.5}{*}{\#Tokens/Video} & \multicolumn{4}{c}{Video-MME (w/o sub.)} \\
        \cmidrule(lr){5-8}
        & & & & Short & Medium & Long & Overall \\
        \midrule
        Baseline (VILA-1.5) & 8 & 1$\times$ & 2048 (=16\textsuperscript{2}$\times$8) & 65.4 & 53.8 & 47.7 & 55.7 \\
        \midrule
        \textbf{Scale} & 32 & 1$\times$ & 8192 (=16\textsuperscript{2}$\times$32) & 73.2 & 58.9 & 50.9 & 61.0 \\
        \textbf{Scale} + \textbf{Compress} & 32 & 4$\times$ & 2048 (=16\textsuperscript{2}$\times$32/4) & 73.7 & 56.7 & 50.0 & 60.1 \\
        \midrule
        \textbf{Scale} + \textbf{Compress} & 256 & 8$\times$ & 8192 (=16\textsuperscript{2}$\times$256/8) & \textbf{75.0} & \textbf{62.2} & \textbf{54.8} & \textbf{64.0} \\
        \bottomrule
    \end{tabular}
\end{table*}

\subsubsection{Spatial ``Scale-Then-Compress''}

For spatial scaling, naively increasing the encoder's input resolution (\eg, to 896$\times$896) imposes uniform overhead on all images regardless of their content. Instead, we adopt \stwo~\cite{shi2024s2}, which extracts multi-scale high-resolution features via adaptive image tiling. Given a vision encoder pre-trained at 448$^2$ resolution and an input image of arbitrary size, \stwo first resizes the image to multiple scales (\eg, 448$^2$, 896$^2$, 1344$^2$); at each scale, it splits the image into 448$^2$ tiles that are independently processed by the encoder. The per-tile feature maps from the same scale are stitched back into a feature map for the whole image at that scale. Finally, feature maps from different scales are interpolated to a common spatial size and concatenated along the channel dimension.

\stwo always resizes images to a square regardless of their original aspect ratio, which can cause distortion for images with extreme aspect ratios. To address this, we propose \textit{\dynstwo}, which adaptively processes images of varying aspect ratios. \dynstwo follows \stwo, but at the largest scale, instead of resizing to a square, it picks the closest size that preserves the input's aspect ratio and is divisible into 448$^2$ tiles. This design is inspired by the dynamic resolution strategy in InternVL~\cite{chen2024internvl15}.

With \dynstwo, NVILA captures richer high-resolution detail, yielding up to \textbf{30 points} of absolute accuracy gain on text-heavy benchmarks (\tab{tab:compression:spatial}). The remaining challenge is to compress these spatial tokens. VILA~\cite{lin2024vila} shows that a simple 2$\times$2 spatial-to-channel (STC) reshape reduces the token count by 4$\times$ without sacrificing accuracy. Pushing further, however, hurts accuracy: increasing the STC ratio to 3$\times$3 leads to a nearly 10-point accuracy drop on DocVQA. We hypothesize that \textit{more aggressive token reductions make the projector substantially harder to train}. To address this, we introduce an additional visual encoder pre-training stage that jointly tunes the vision encoder and projector. This stage recovers most of the accuracy loss from 3$\times$3 compression while preserving its \textbf{2.4$\times$} speedup over the 2$\times$2 baseline in both training and inference.

We also explored alternative designs for spatial token compression, such as TokenLearner from RT-1~\cite{brohan2022rt1} and Perceiver Resampler from MiniCPM-V~\cite{yao2024minicpmv}. At the same token-reduction ratio, these learnable methods perform no better than the simple spatial-to-channel design, even with the additional visual encoder pre-training stage. We attribute this to optimization difficulty rather than representational capacity, and leave a deeper investigation to future work.

\subsubsection{Temporal ``Scale-Then-Compress''}

For temporal scaling, we uniformly sample more frames per video. Following prior work~\cite{xue2024longvila}, we include additional video SFT to enable the model to process longer sequences. As shown in \tab{tab:compression:temporal}, increasing from 8 to 32 frames improves Video-MME accuracy by \textbf{more than 5 points}, but it also increases the number of visual tokens by 4$\times$.

As with spatial tokens, we compress the temporal representations to recover efficiency. Since consecutive frames often contain similar information, we adopt \textit{temporal averaging}~\cite{wang2016tsn}, which partitions frames into groups and pools visual tokens within each group. This reduces temporal redundancy while retaining important spatiotemporal detail. Empirically, 4$\times$ compression yields only a modest accuracy drop. Compared with the original baseline at the same token budget, the scale-then-compress model has nearly the same cost\footnote{Running the visual encoder on more frames adds overhead, but this is not the runtime bottleneck.} and substantially higher accuracy. Pushing further to 256 frames with an 8$\times$ compression ratio yields state-of-the-art results among 7--8B open-source models on Video-MME (\tab{tab:results:video}).

\subsection{Efficient Training}

Training a state-of-the-art VLM is costly and compute-intensive. This section explores system-algorithm co-design for efficient VLM training. On the algorithm side, we examine an unsupervised dataset pruning method that streamlines the training set. On the system side, we investigate FP8 mixed-precision training for acceleration.

\begin{figure*}[t]
    \centering
    \includegraphics[width=1.0\linewidth]{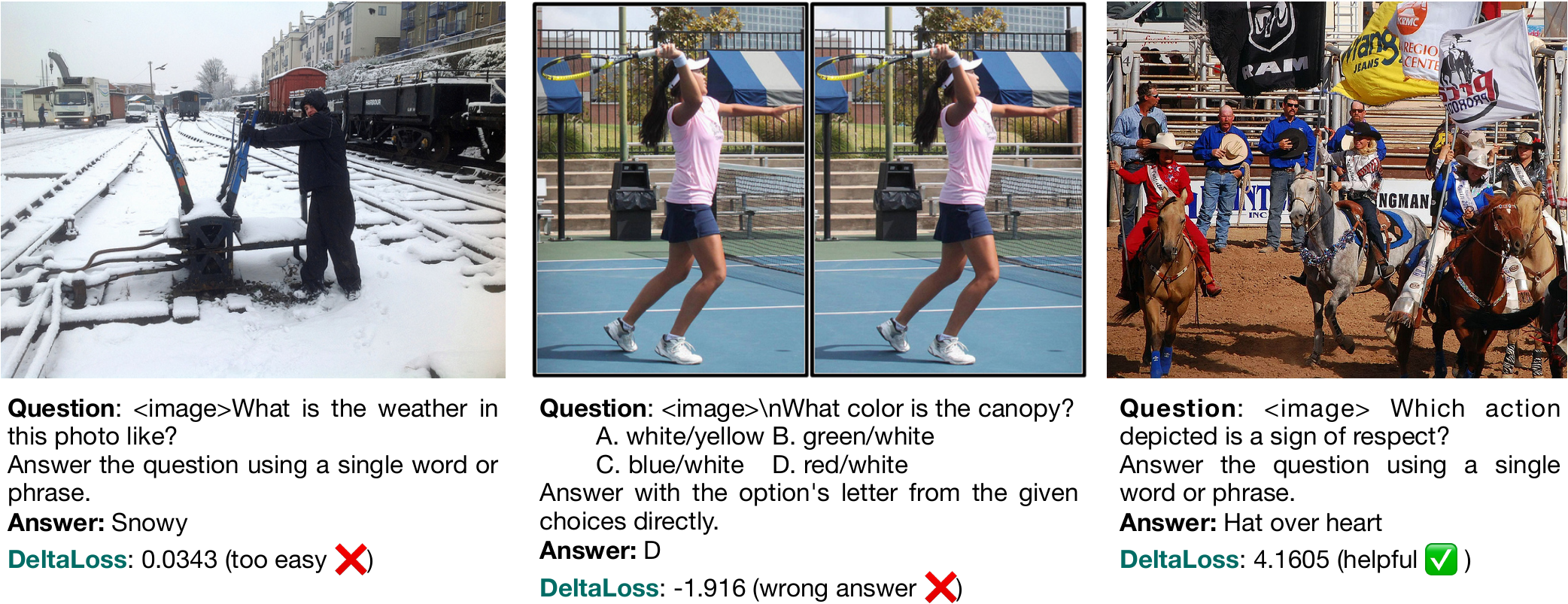}
    \caption{\textbf{DeltaLoss visualization}. \textit{Left}, \textit{Middle}, and \textit{Right} panels show training examples scored as too easy, distracting, and informative by DeltaLoss, respectively.}
    \label{fig:deltaloss_examples}
\end{figure*}

\subsubsection{Dataset Pruning}

Prior work~\cite{tong2024cambrian,li2024llavaonevision,laurencon2024idefics2} has aggregated ever-larger SFT datasets from diverse sources, yielding consistent benchmark improvements. However, \textit{not all data contributes equally}, and unchecked dataset growth leads to significant redundancy. In NVILA, we apply the ``\textit{scale-then-compress}'' principle to data: we first expand our SFT dataset mixture and then compress it via pruning. While prior work has explored data selection for vision inputs~\cite{coleman2019svp,xu2023metaclip,abbas2023semdedup} and text-only inputs~\cite{tirumala2023d4,xia2024less,gu2024pds}, few studies address the mixed image-text setting of VLM training. With tens of millions of training samples, pruning the dataset without sacrificing accuracy is essential.

Inspired by recent work in knowledge distillation~\cite{gu2024miniplm}, we score each training example by the \textit{DeltaLoss}
\begin{equation}
    \Delta(x) \;=\; \log\frac{p_{\text{large}}(x)}{p_{\text{small}}(x)},
    \label{eq:deltaloss}
\end{equation}
where $p_{\text{large}}(x)$ and $p_{\text{small}}(x)$ are the probabilities a large and a small reference VLM, respectively, assign to the answer tokens of $x$. Given the full SFT mixture $D=\bigcup_{i=1}^{N} D_i$ partitioned into $N$ source-level subsets and a target keep-ratio $\rho\in(0,1]$, we form the pruned training set
\begin{equation}
    D' \;=\; \bigcup_{i=1}^{N} \mathrm{top}_{\,\lceil \rho\,|D_i|\rceil}\!\left\{\,\Delta(x) \;\big|\; x\in D_i\,\right\},
\end{equation}
which keeps the top-scoring $\lceil \rho\,|D_i|\rceil$ examples within each subset. The main motivation is to \textbf{filter out examples that are either too easy or too distracting}:
\begin{compactitem}
\item If both models answer correctly, or both fail, $\Delta(x)$ is near zero; the example offers little discriminative signal.
\item When the small model answers correctly but the large model fails, $\Delta(x)$ is negative, suggesting the example tends to distract learning and will eventually be forgotten by a more capable model.
\item When the small model fails but the large model succeeds, $\Delta(x)$ is positive, indicating strong supervision: challenging for small models but learnable by larger ones.
\end{compactitem}
We apply DeltaLoss to each subset $D_i$ and ablate over a range of keep-ratios $\rho$.

\begin{table}[t]
    \setlength{\tabcolsep}{4pt}
    \renewcommand{\arraystretch}{1.1}
    \footnotesize\centering
    \caption{\textbf{Dataset pruning on the NVILA recipe}. DeltaLoss consistently outperforms other data selection methods, with minimal accuracy loss at a 50\% keep-ratio.}
    \label{tab:dataset_prune}
    \begin{tabular}{lcccc}
        \toprule
         \textbf{Method} & IM-10 & MMMU & DocVQA & TextVQA \\
        \midrule
        100\% (baseline)     & 75.6  & 48.0 & 90.1 & 78.8 \\
        \midrule
        \rowcolor{black!10} \multicolumn{5}{l}{50\%} \\
        DeltaLoss~\cite{gu2024miniplm}      & \textbf{75.5}  & \textbf{48.1} & \textbf{89.7} & \textbf{78.4} \\
        Cluster Pruning      &  74.5 & 47.8 & 88.3  & 77.0 \\
        Random  Pruning      &  74.0 & 47.6 & 87.1 & 76.6 \\
        \midrule
        \rowcolor{black!10} \multicolumn{5}{l}{30\%} \\
        DeltaLoss~\cite{gu2024miniplm}     & \textbf{74.0}  & \textbf{47.8}  & \textbf{87.9} & \textbf{76.4}     \\
        Cluster Pruning   &  73.5  &  47.7  &  84.1  &  76.0\\
        Random  Pruning   &  73.1  &  47.7  &  82.9  &  75.6 \\
        \midrule
        \rowcolor{black!10} \multicolumn{5}{l}{10\%} \\
        DeltaLoss~\cite{gu2024miniplm}      & \textbf{72.4}  & {47.1}  & \textbf{84.4} & \textbf{74.5}    \\
        Cluster Pruning      &  72.2 & \textbf{47.4} & 79.6 & 73.2 \\
        Random  Pruning      & 72.0  & 47.0 & 77.3 & 72.6  \\
        \bottomrule
    \end{tabular}
\end{table}

\begin{table}[t]
    \setlength{\tabcolsep}{4pt}
    \renewcommand{\arraystretch}{1.1}
    \footnotesize\centering
    \caption{\textbf{Dataset pruning on the Pixmo recipe~\cite{deitke2024molmo}}. DeltaLoss outperforms other data selection methods in most settings, with negligible accuracy drop at a 50\% keep-ratio.}
    \label{tab:dataset_prune_pixmo}
    \begin{tabular}{lcccc}
        \toprule
         \textbf{Method} & IM-10 & MMMU & DocVQA & TextVQA \\
        \midrule
        100\% (baseline) & 74.9 & 45.8 & 90.0 & 76.4 \\
        \midrule
        \rowcolor{black!10} \multicolumn{5}{l}{50\%} \\
        DeltaLoss~\cite{gu2024miniplm}  & \textbf{75.2} & \textbf{47.2} & 89.2 & \textbf{76.5} \\
        Cluster Pruning  & 75.0 & 47.1 & 89.0 & 76.3 \\
        Random Pruning   & 74.6 & 45.1 & \textbf{89.6} & 76.2 \\
        \midrule
        \rowcolor{black!10} \multicolumn{5}{l}{30\%} \\
        DeltaLoss~\cite{gu2024miniplm}  & \textbf{74.3} & \textbf{46.8} & \textbf{88.7} & \textbf{74.7} \\
        Cluster Pruning   & 73.4 & 46.4 & 87.2 & 74.6 \\
        Random Pruning    & 72.7 & 45.4 & 88.8 & 74.1 \\
        \midrule
        \rowcolor{black!10} \multicolumn{5}{l}{10\%} \\
        DeltaLoss~\cite{gu2024miniplm} & \textbf{72.6} & \textbf{46.5} & \textbf{88.1} & \textbf{74.2} \\
        Cluster Pruning   & 72.2 & 46.2 & 87.5 & 73.9 \\
        Random Pruning    & 71.4 & 45.4 & 87.7 & 72.8 \\
        \bottomrule
    \end{tabular}
\end{table}

We evaluate DeltaLoss against cluster pruning and random pruning (\tab{tab:dataset_prune}). For random pruning, we sample examples uniformly at random and report the mean over three runs. For cluster pruning, we apply $k$-means clustering on SigLIP features and prune uniformly across clusters. We sweep three keep-ratios ($\rho\in\{10\%, 30\%, 50\%\}$) and report average performance across 10 image benchmarks (IM-10). DeltaLoss consistently outperforms both baselines; on DocVQA in particular, random pruning degrades sharply while DeltaLoss retains accuracy. At a 50\% keep-ratio, the average score remains competitive while training time is halved. We therefore adopt $\rho=50\%$ for all subsequent experiments.

We further examine the impact of data pruning on newly added datasets. We incorporate varying portions of Pixmo data~\cite{deitke2024molmo} into the NVILA training set. As shown in \tab{tab:dataset_prune_pixmo}, naively combining Pixmo with the NVILA training set degrades DocVQA and TextVQA while only improving MMMU, suggesting that indiscriminate dataset growth can hurt performance. Applying DeltaLoss to prune the Pixmo data yields consistent improvements across benchmarks, even at small keep-ratios.

\subsubsection{FP8 Training}

\begin{table}[t]
    \setlength{\tabcolsep}{5pt}
    \renewcommand{\arraystretch}{1.1}
    \footnotesize\centering
    \caption{\textbf{FP8 training}. FP8 accelerates NVILA training without degrading accuracy, with the largest gains when gradient checkpointing (GC) is disabled. Throughput is measured at the largest batch size (BS) that fits on 64 H100 GPUs. Video-MME uses an 8-frame setting with subtitles.}
    \label{tab:analysis:fp8}
    \begin{tabular}{lccccc}
        \toprule
         & GC & BS & Throughput & MMMU & Video-MME \\
        \midrule
        BF16 & \xmark & 4 & 199.2 (1.0$\times$) & 47.9 & 52.9 \\ 
        FP8 & \xmark & 16 & 390.1 (2.0$\times$) & 47.0 & 53.0 \\     
        \midrule
        BF16 & \cmark & 30  & 491.7 (2.5$\times$) & 47.8 & 53.1 \\ 
        FP8 & \cmark & 36 & 579.9 (2.9$\times$) & 47.7 & 53.0 \\
        \bottomrule
    \end{tabular}
\end{table}

FP16~\cite{micikevicius2018mixed} and BF16~\cite{kalamkar2019bf16} are the standard training precisions, trading off some of FP32's numerical stability for faster compute on NVIDIA GPUs. The NVIDIA Hopper and Blackwell architectures (\eg, H100 and B200) now also support FP8 natively, promising greater compute and memory efficiency than BF16.

FP8 has been widely applied to LLM training. NVIDIA's Transformer Engine performs matrix multiplications (GEMM) in FP8, accelerating training. FP8-LM~\cite{peng2023fp8lm} additionally quantizes gradients, the weight master copy, and the first-order momentum to FP8, reducing communication overhead and memory footprint. COAT~\cite{xi2024coat} further compresses activations and the optimizer's second-order momentum, improving memory efficiency while maintaining accuracy.

We adopt the FP8 implementation from COAT~\cite{xi2024coat} to accelerate NVILA training. A key difference between LLM and VLM training is sequence-length variability. After packing, LLM training samples tend to be similar in length, so throughput is relatively insensitive to batch size. VLM training samples, in contrast, vary widely in length: video samples may require tens of thousands of tokens, image samples a few hundred to a few thousand, and text-only samples far fewer. As a result, batches dominated by short samples underutilize the GPU and benefit substantially from a larger batch size. As shown in \tab{tab:analysis:fp8}, applying FP8 to both weights and activations allows NVILA to increase the batch size from 4 to 16, producing a 2$\times$ speedup. When gradient checkpointing is enabled, quantizing activations is less essential; instead, we integrate the cross-entropy kernel from Liger~\cite{hsu2024liger} to reduce peak memory due to Qwen's large vocabulary. In this setting, FP8 training still provides a 1.2$\times$ speedup over BF16.

\begin{table}[t]
    \setlength{\tabcolsep}{2.25pt}
    \renewcommand{\arraystretch}{1.1}
    \footnotesize\centering
    \caption{\textbf{Fine-tuning recipe}. Our recommendation is to tune the LLM with either LoRA or QLoRA and to tune ViT's layer normalization (LN) layers with a much smaller learning rate. This setup achieves competitive accuracy and is also the most memory- and compute-efficient. All experiments use a batch size of 1 with gradient checkpointing disabled, and throughput is measured on a single NVIDIA A100 80GB GPU. For settings with \{1,5,10,50\}, we select the learning rate ratio from this set that gives the best results for each benchmark. ``\textit{FT-5}'' refers to the average accuracy across AITZ~\cite{zhang2024aitz}, ALFRED~\cite{shridhar2020alfred}, nuScenes~\cite{caesar2020nuscenes}, PathVQA~\cite{he2020pathvqa}, and Widget Caption~\cite{li2020widgetcaption}.}
    \begin{tabular}{llcccc}
        \toprule
        ViT & LLM & \makecell{Memory \\ (GB)} & \makecell{Throughput \\ (iter/s)} & $\texttt{LR}_\texttt{LLM}/\texttt{LR}_\texttt{ViT}$ & \makecell{Accuracy \\ (FT-5)} \\
        \midrule
        \multirow{2}{*}{LoRA} & \multirow{2}{*}{LoRA} & \multirow{2}{*}{20.1} & \multirow{2}{*}{3.4} & 1 & 69.2 \\
        & & & & \{1,5,10,50\} & \textbf{71.8} \\
        \midrule
        \multirow{2}{*}{LN} & \multirow{2}{*}{LoRA} & \multirow{2}{*}{19.2} & \multirow{2}{*}{4.5} & 1 & 63.5 \\
        & & & & \{1,5,10,50\} & \textbf{71.4} \\
        \midrule
        \multirow{2}{*}{FT} & \multirow{2}{*}{LoRA} & \multirow{2}{*}{21.9} & \multirow{2}{*}{4.2} & 1 & 64.0 \\
        & & & & \{1,5,10,50\} & \textbf{70.1} \\
        \midrule
        \multirow{2}{*}{LoRA} & \multirow{2}{*}{QLoRA} & \multirow{2}{*}{11.1} & \multirow{2}{*}{2.6} & 1 & 63.0 \\
        & & & & \{1,5,10,50\} & \textbf{70.8} \\
        \midrule
        \multirow{2}{*}{LN} & \multirow{2}{*}{QLoRA} & \multirow{2}{*}{10.2} & \multirow{2}{*}{3.1} & 1 & 62.7 \\
        & & & & \{1,5,10,50\} & \textbf{70.9} \\
        \midrule
        \textcolor{black!50}{FT} & \textcolor{black!50}{FT} & \textcolor{black!50}{63.5} & \textcolor{black!50}{6.1} & \textcolor{black!50}{1} & \textcolor{black!50}{77.7} \\
        \bottomrule
    \end{tabular}
    \label{tab:finetune}
\end{table}

\subsection{Efficient Fine-Tuning}

Once a foundation VLM is trained, domain-specific fine-tuning adapts it to specialized tasks. Conventional parameter-efficient fine-tuning (PEFT) methods are largely studied in the context of LLMs, and how to best fine-tune a VLM remains underexplored. In NVILA, we find that (i) the learning rate should be set differently for the ViT and the LLM, and (ii) tuning only a small subset of ViT parameters can match full PEFT.

When jointly fine-tuning the vision encoder (ViT) and the language model (LLM) with PEFT methods, we observe that the ViT's learning rate should be 5--50$\times$ smaller than the LLM's. We also find that fine-tuning only the ViT's LayerNorm layers matches the accuracy of LoRA (\tab{tab:finetune}) while reducing training time by 25\% relative to applying LoRA to the vision encoder. With this configuration, NVILA can be fine-tuned to various downstream tasks within 24\,GB of GPU memory while maintaining on-par performance.

\begin{table}[t]
    \setlength{\tabcolsep}{4pt}
    \renewcommand{\arraystretch}{1.1}
    \footnotesize\centering
    \caption{\textbf{Quantization recipe}. W4A16 quantization on the LLM backbone introduces a small accuracy drop, while W8A8 on the ViT is nearly lossless. Together, they reduce TTFT by 28\%. TTFT: time-to-first-token.}
    \label{tab:analysis:quantization}
    \begin{tabular}{cccccc}
        \toprule
        ViT & LLM & AI2D  & MMMU & VideoMME  & TTFT (s)\\
        \midrule
        FP16 & FP16 & 91.0 & 50.7 &  63.9  & 0.90\\
        \midrule
        FP16 & W4A16 & 90.9 & 49.2 &   62.0 & 0.77\\
        W8A8 & W4A16 & 90.9  & 49.3	& 62.1  & 0.65 \\
        \bottomrule
    \end{tabular}
\end{table}

\subsection{Efficient Deployment}

VLMs are increasingly deployed in resource-constrained edge settings such as robotics, where both latency and memory are limited. To deploy NVILA efficiently, we build a specialized inference engine that targets the two phases of inference, prefilling and decoding, with phase-specific quantization.

In the compute-bound prefilling stage, we first apply token compression (\sect{sec:approach:efficient_arch}) to reduce the LLM backbone's inference workload. After compression, the vision tower becomes the primary bottleneck, accounting for over 90\% of the prefilling latency. We therefore apply W8A8 quantization to the vision tower to reduce NVILA's time-to-first-token (TTFT).
For the memory-bound decoding stage, we follow AWQ~\cite{lin2024awq} and apply W4A16 quantization to the LLM backbone. We further optimize the original AWQ implementation by introducing FP16 accumulation in the W4A16 GEMM kernels, yielding a 1.7$\times$ kernel speedup without compromising accuracy. A detailed comparison is provided in \fig{fig:inference_efficiency}.

\section{Experiments}
\label{sec:exp}

\subsection{Training Details}

\begin{table}[t]
    \footnotesize\centering
    \setlength{\tabcolsep}{2pt}
    \renewcommand{\arraystretch}{1.1}
    \caption{\textbf{Training recipe}. NVILA adds two stages to VILA's pipeline: Stage 2 pre-trains the visual encoder to recover the accuracy lost to spatial token compression, and Stage 5 fine-tunes the model on video data to extend long-video understanding. ``LR'' denotes the initial learning rate of the trainable components in each stage.}
    \label{tab:recipe:training}
    \begin{tabular}{lcccc}
        \toprule
         & Visual Encoder & Projector & \multicolumn{2}{c}{Token Processor} \\
         \cmidrule(lr){4-5}
         & (ViT) & (MLP) & (LLM) & LR \\
        \midrule
        Initial & from \cite{zhai2023siglip} & random & from \cite{yang2024qwen2} & -- \\ 
        \midrule
        Stage 1 & \textcolor{black!25}{\textit{frozen}} & trainable & \textcolor{black!25}{\textit{frozen}} & 1$\times$10\textsuperscript{-3} \\
        Stage 2 & trainable & trainable & \textcolor{black!25}{\textit{frozen}} & 5$\times$10\textsuperscript{-5} \\
        Stage 3 & \textcolor{black!25}{\textit{frozen}} & trainable & trainable & 5$\times$10\textsuperscript{-5} \\
        Stage 4 & trainable & trainable & trainable & 2$\times$10\textsuperscript{-5} \\
        Stage 5 & trainable & trainable & trainable & 2$\times$10\textsuperscript{-5} \\
        \bottomrule
    \end{tabular}
\end{table}

We train NVILA in a five-stage pipeline: (1) \textit{projector initialization}, (2) \textit{visual encoder pre-training}, (3) \textit{token processor pre-training}, (4) \textit{image instruction-tuning}, and (5) \textit{video instruction-tuning}. Stages 1, 3, and 4 are also part of VILA training. Stage 2 recovers the accuracy loss caused by spatial token compression (\tab{tab:compression:spatial}), and Stage 5 extends the model's long-video understanding capability. The detailed training recipe is given in \tab{tab:recipe:training}, and the data recipe in \tab{tab:recipe:data}.

Our implementation is built on PyTorch 2.3.0~\cite{paszke2019pytorch,ansel2024pytorch2} and Transformers 4.46.0~\cite{wolf2020transformers}. We use DeepSpeed 0.9.5~\cite{rasley2020deepspeed} to shard large models across devices and gradient checkpointing to reduce memory usage. We adopt FlashAttention-2~\cite{dao2024flashattention2} to accelerate training in both the LLM and the visual encoder. We also implement function-preserving, on-the-fly sequence packing to fuse samples of different lengths, yielding roughly a 30\% speedup. All models are trained on 128 NVIDIA H100 GPUs with a global batch size of 2048 across all stages, using AdamW without weight decay. We adopt a cosine learning-rate schedule with a linear warm-up over the first 3\% of training. Per-stage initial learning rates are listed in \tab{tab:recipe:training}.

We release two model variants: \textbf{NVILA-Lite} maximizes efficiency by applying all the techniques described in this paper, while \textbf{NVILA} trades a small amount of efficiency for higher accuracy. Both variants share the same training pipeline.

\subsection{Accuracy Results}

\begin{table*}[t]
    \setlength{\tabcolsep}{2pt}
    \renewcommand{\arraystretch}{1.1}
    \footnotesize\centering
    \caption{\textbf{Image benchmarks}. Best result among open-source models within each size group is in \textbf{bold} and the second-best is \underline{underlined}; proprietary and grayed-out larger models ($>$15B parameters) are shown for reference and are not included in bolding.}
    \begin{tabular}{lrcccccccccccc}
        \toprule
        & & AI2D & ChartQA & DocVQA & InfoVQA & MathVista & \multicolumn{3}{c}{MMMU} & \multirow{2.5}{*}{\makecell{Real \\ WorldQA}} & SEED & TextVQA & VQAv2 \\
        \cmidrule(lr){3-3}\cmidrule(lr){4-4}\cmidrule(lr){5-5}\cmidrule(lr){6-6}\cmidrule(lr){7-7}\cmidrule(lr){8-10}\cmidrule(lr){12-12}\cmidrule(lr){13-13}\cmidrule(lr){14-14}
        & & test & test & test & test & testmini & val & test & pro & & image & val & testdev \\
        \midrule
        GPT-4o & -- & 94.2 & 85.7 & 92.8 & 79.2 & 63.8 & 69.1 & 64.7 & 51.9 & 75.4 & 76.2 & 77.4 & 78.7 \\
        Claude 3.5 Sonnet & -- & 94.7 & 90.8 & 85.2 & 74.3 & 67.7 & 68.3 & 63.7 & 51.5 & 60.1 & -- & 74.1 & 70.7 \\
        Gemini 1.5 Pro & -- & 94.4 & 87.2 & 93.1 & 81.0 & 63.9 & 62.2 & 57.6 & 43.5 & 70.4 & -- & 78.7 & 80.2 \\
        \midrule
        LLaVA-1.5 & 7B & 55.5 & 17.8 & 28.1 & 25.8 & 25.6 & 35.7 & -- & -- & 54.8 & 66.1 & 58.2 & 78.5 \\
        VILA-1.5 & 8B & 76.6 & 52.7 & 40.6 & 25.9 & 36.7 & 38.6 & 32.7 & -- & 52.7 & 73.8 & 68.5 & 83.0 \\
        Cambrian-1 & 8B & 73.0 & 73.3 & 77.8 & 41.6 & 49.0 & 42.7 & -- & -- & 64.2 & 74.7 & 71.7 & 81.2 \\
        Florence-VL & 8B & 74.2 & 74.7 & 84.9 & 51.7 & 55.5 & 43.7 & -- & -- & 64.2 & 74.9 & 74.2 & 84.7 \\
        LLaVA-OneVision & 8B & 81.4 & 80.0 & 87.5 & 68.8 & 63.2 & 48.8 & 42.8 & 24.1 & 66.3 & 75.4 & 78.3 & 84.0 \\
        Llama 3.2 & 11B & \underline{91.9} & 83.4 & 88.4 & -- & 51.5 & 50.7 & -- & -- & -- & -- & -- & 75.2 \\
        InternVL2 & 8B & 83.8 & 83.3 & 91.6 & \underline{74.8} & 58.3 & \underline{51.2} & 42.6 & \underline{29.0} & 64.2 & 76.2 & 77.4 & 76.7 \\
        Qwen2-VL & 8B & 83.0 & 83.0 & \textbf{94.5} & \textbf{76.5} & 58.2 & \textbf{54.1} & \textbf{46.6} & \textbf{30.5} & \textbf{70.1} & 76.0 & \textbf{84.3} & 82.9 \\
        \textbf{NVILA-Lite} & 8B & 91.0 & \underline{84.8} & 91.7 & 67.9 & \underline{64.5} & 50.7 & \underline{45.7} & 26.5 & 65.6 & \underline{76.3} & 78.1 & \underline{85.0} \\
        \textbf{NVILA} & 8B & \textbf{92.3} & \textbf{86.1} & \underline{93.7} & 70.7 & \textbf{65.4} & 49.9 & 44.4 & 27.8 & \underline{68.6} & \textbf{76.5} & \underline{80.1} & \textbf{85.4} \\
        \midrule
        LLaVA-1.5 & 13B & 61.1 & 18.2 & 30.3 & 29.4 & 27.7 & 37.0 & -- & -- & 55.3 & 68.2 & 61.3 & 80.0 \\
        VILA-1.5 & 13B & 79.9 & 59.5 & 58.6 & 30.4 & 42.7 & 37.9 & \underline{33.6} & -- & 57.5 & 72.6 & 65.0 & 82.8 \\
        Cambrian-1 & 13B & 73.6 & 73.8 & 76.8 & -- & 48.0 & 40.0 & -- & -- & 63.0 & 74.4 & 72.8 & -- \\
        Pixtral & 12B & 79.0 & \underline{81.8} & \underline{90.7} & 50.8 & 58.0 & 52.5 & -- & -- & 65.4 & -- & 75.7 & 80.2 \\
        \textbf{NVILA-Lite} & 15B & \underline{92.0} & \underline{81.8} & 90.6 & \underline{69.3} & \underline{61.7} & \textbf{58.7} & \textbf{51.8} & \underline{33.7} & \underline{67.1} & \underline{75.6} & \underline{77.3} & \underline{83.7} \\
        \textbf{NVILA} & 15B & \textbf{94.1} & \textbf{86.9} & \textbf{94.0} & \textbf{73.5} & \textbf{66.1} & \underline{56.7} & \textbf{51.8} & \textbf{33.8} & \textbf{69.5} & \textbf{76.6} & \textbf{80.0} & \textbf{84.8} \\
        \midrule
        \textcolor{black!50}{LLaVA-NeXT} & \textcolor{black!50}{34B} & \textcolor{black!50}{--} & \textcolor{black!50}{--} & \textcolor{black!50}{--} & \textcolor{black!50}{--} & \textcolor{black!50}{46.5} & \textcolor{black!50}{48.1} & \textcolor{black!50}{44.5} & \textcolor{black!50}{22.9} & \textcolor{black!50}{--} & \textcolor{black!50}{75.9} & \textcolor{black!50}{69.5} & \textcolor{black!50}{83.7} \\
        \textcolor{black!50}{Cambrian-1} & \textcolor{black!50}{34B} & \textcolor{black!50}{79.7} & \textcolor{black!50}{75.6} & \textcolor{black!50}{75.5} & \textcolor{black!50}{46.0} & \textcolor{black!50}{53.2} & \textcolor{black!50}{49.7} & \textcolor{black!50}{--} & \textcolor{black!50}{--} & \textcolor{black!50}{67.8} & \textcolor{black!50}{75.3} & \textcolor{black!50}{76.7} & \textcolor{black!50}{83.8} \\
        \textcolor{black!50}{VILA-1.5} & \textcolor{black!50}{40B} & \textcolor{black!50}{88.9} & \textcolor{black!50}{67.8} & \textcolor{black!50}{58.6} & \textcolor{black!50}{38.4} & \textcolor{black!50}{49.3} & \textcolor{black!50}{51.9} & \textcolor{black!50}{46.9} & \textcolor{black!50}{25.0} & \textcolor{black!50}{60.8} & \textcolor{black!50}{69.1} & \textcolor{black!50}{73.6} & \textcolor{black!50}{84.3} \\
        \textcolor{black!50}{InternVL2} & \textcolor{black!50}{40B} & \textcolor{black!50}{87.1} & \textcolor{black!50}{86.2} & \textcolor{black!50}{93.9} & \textcolor{black!50}{78.7} & \textcolor{black!50}{63.7} & \textcolor{black!50}{55.2} & \textcolor{black!50}{47.4} & \textcolor{black!50}{34.2} & \textcolor{black!50}{71.8} & \textcolor{black!50}{78.2} & \textcolor{black!50}{83.0} & \textcolor{black!50}{--} \\
        \textcolor{black!50}{LLaVA-OneVision} & \textcolor{black!50}{72B} & \textcolor{black!50}{85.6} & \textcolor{black!50}{83.7} & \textcolor{black!50}{91.3} & \textcolor{black!50}{74.9} & \textcolor{black!50}{67.5} & \textcolor{black!50}{56.8} & \textcolor{black!50}{52.3} & \textcolor{black!50}{31.0} & \textcolor{black!50}{71.9} & \textcolor{black!50}{75.4} & \textcolor{black!50}{80.5} & \textcolor{black!50}{85.2} \\
        \textcolor{black!50}{NVLM-D-1.0} & \textcolor{black!50}{78B} & \textcolor{black!50}{94.2} & \textcolor{black!50}{86.0} & \textcolor{black!50}{92.6} & \textcolor{black!50}{--} & \textcolor{black!50}{65.2} & \textcolor{black!50}{59.7} & \textcolor{black!50}{54.6} & \textcolor{black!50}{--} & \textcolor{black!50}{69.7} & \textcolor{black!50}{--} & \textcolor{black!50}{82.1} & \textcolor{black!50}{85.4} \\
        \textcolor{black!50}{Llama 3.2} & \textcolor{black!50}{90B} & \textcolor{black!50}{92.3} & \textcolor{black!50}{85.5} & \textcolor{black!50}{90.1} & \textcolor{black!50}{--} & \textcolor{black!50}{57.3} & \textcolor{black!50}{60.3} & \textcolor{black!50}{--} & \textcolor{black!50}{39.5} & \textcolor{black!50}{--} & \textcolor{black!50}{--} & \textcolor{black!50}{--} & \textcolor{black!50}{--} \\
        \bottomrule
    \end{tabular}
    \label{tab:results:image}
\end{table*}

\subsubsection{Image Benchmarks}

We conduct comprehensive evaluations on a diverse set of image benchmarks (\tab{tab:results:image}): AI2D~\cite{kembhavi2016diagram}, ChartQA~\cite{masry2022chartqa}, DocVQA~\cite{mathew2021docvqa}, InfographicVQA~\cite{mathew2022infographicvqa}, MathVista~\cite{lu2024mathvista}, MMMU~\cite{yue2024mmmu} (with zero-shot CoT), RealWorldQA~\cite{xai2024grok15}, SEED-Bench~\cite{li2024seed}, TextVQA~\cite{singh2019towards}, and VQAv2~\cite{goyal2017making}.

NVILA is competitive with the strongest open-source models in each size category, including Qwen2-VL~\cite{wang2024qwen2vl}, InternVL2~\cite{chen2024internvl}, and Pixtral~\cite{agrawal2024pixtral}.
On general VQA tasks (ChartQA, DocVQA, InfoVQA, TextVQA, VQAv2, SEED), NVILA (8B and 15B) is on par with or surpasses proprietary models such as GPT-4o and Gemini~1.5~Pro.
On the science benchmark AI2D, NVILA leads its size group at both 8B and 15B, with NVILA-15B (94.1) approaching the proprietary frontier (GPT-4o 94.2, Gemini~1.5~Pro 94.4, Claude~3.5~Sonnet 94.7).
On reasoning and knowledge benchmarks (MMMU, RealWorldQA, MathVista), NVILA-15B leads among open-source models in its size group, with the largest gain over the 8B model coming on MMMU.
On OCR-heavy benchmarks (ChartQA, DocVQA, TextVQA), the 8B model is also highly competitive.
We provide qualitative examples in \fig{fig:qualitative_examples} to illustrate NVILA's OCR, reasoning, and multi-image capabilities.

\subsubsection{Video Benchmarks}

We evaluate our models on a range of video understanding benchmarks~\cite{yu2019activitynetqa,zhou2024mlvu,li2024mvbench,fu2024videomme}, spanning short clips of a few seconds to videos up to an hour in duration. \tab{tab:results:video} compares NVILA against strong baselines~\cite{shu2024videoxl,shen2024longvu,wang2024qwen2vl,li2024llavaonevision,liu2024oryx,xue2024longvila}.
With the ``\textit{scale-then-compress}'' design, NVILA supports long-context video understanding with up to 256 frames. NVILA-8B achieves state-of-the-art results on all evaluated benchmarks among open-source 7--8B models and matches the proprietary GPT-4o mini.

\begin{table*}[t]
    \setlength{\tabcolsep}{4pt}
    \renewcommand{\arraystretch}{1.1}
    \footnotesize\centering
    \caption{\textbf{Video benchmarks}. Best result in \textbf{bold}. \#F: number of frames sampled. acc.: accuracy. m-avg: M-Avg metric (multiple-choice average) used by MLVU. mc: multiple-choice. w/o\,sub.\,/\,w/\,sub.: without\,/\,with subtitles.}
    \begin{tabular}{lrcccccccccc}
        \toprule
        & & & \multicolumn{2}{c}{ActivityNet-QA} & \multicolumn{2}{c}{LongVideoBench} & MLVU & MVBench & NExT-QA & \multicolumn{2}{c}{Video-MME} \\
        \cmidrule(lr){4-5}\cmidrule(lr){6-7}\cmidrule(lr){8-8}\cmidrule(lr){9-9}\cmidrule(lr){10-10}\cmidrule(lr){11-12}
        & & \#F & acc. & score & val & test & m-avg & test & mc & w/o sub. & w/ sub. \\
        \midrule
        GPT-4o mini & -- & -- & -- & -- & 56.5 & 58.8 & -- & -- & -- & 64.8 & 68.9\\
        GPT-4o & -- & -- & 61.9 & -- & 66.7 & 66.7 & 64.6 & -- & -- & 71.9 & 77.2 \\
        \midrule
        LLaVA-NeXT-Video & 7B & 32 & 53.5 & 3.2 & 43.5 & 43.5 & -- & 33.7 & -- & 46.5 & -- \\
        Video-XL & 7B & 2048 & -- & -- & 49.5 & 51.3 & 64.9 & 55.3 & 77.2 & 55.5 & 61.0 \\
        InternVL2 & 8B & 64 & -- & -- & 54.6 & -- & 64.0 & 65.8 & -- & 56.3 & 59.3 \\
        LLaVA-OneVision & 8B & 32 & 56.6 & -- & 56.5 & -- & 64.7 & 56.7 & 79.4 & 58.2 & 61.5\\
        Oryx-1.5 & 8B & 128 & -- & -- & 56.3 & -- & 67.5 & 67.6 & 81.8 & 58.8 & 64.2 \\
        LongVILA & 7B & 256 & 59.5 & -- & 57.1 & -- & -- & 67.1 & 80.7 & 60.1 & 65.1 \\
        LongVU & 7B & 1fps & -- & -- & -- & -- & 65.4 & 66.9 & -- & 60.6 & -- \\
        Qwen2-VL & 8B & 2fps & -- & -- & 55.6 & 56.8 & 65.5 & 67.0 & -- & 63.3 & 69.0\\
        \textbf{NVILA} & 8B & 256 & \textbf{60.9} & \textbf{3.7} & \textbf{57.7} & \textbf{58.7} & \textbf{70.1} & \textbf{68.1} & \textbf{82.2} & \textbf{64.2} & \textbf{70.0} \\
        \bottomrule
    \end{tabular}
    \label{tab:results:video}
\end{table*}

\subsection{Efficiency Results}

\begin{figure*}[t]
    \centering
    \includegraphics[width=\linewidth]{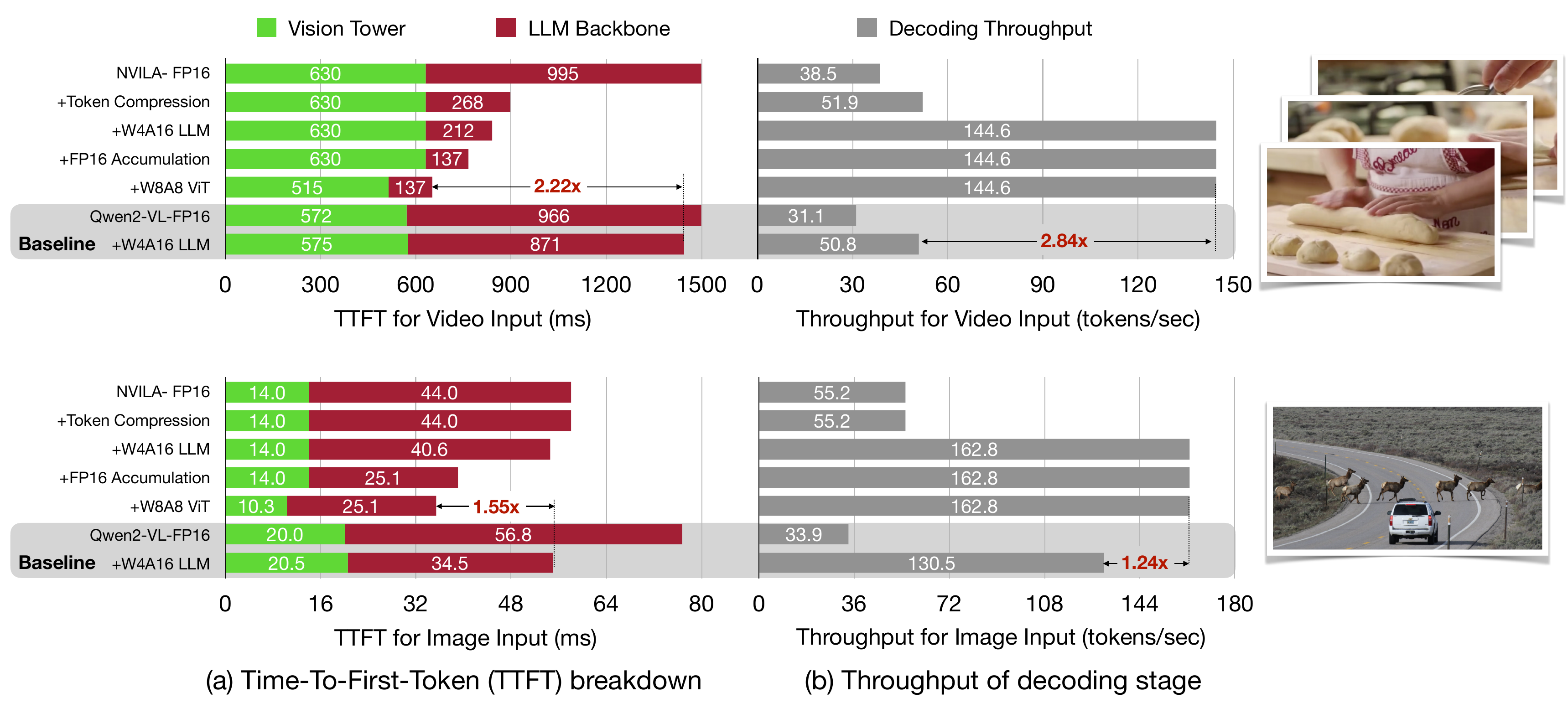}
    \caption{\textbf{Inference efficiency breakdown}. We benchmark NVILA-8B against Qwen2-VL-7B~\cite{wang2024qwen2vl} on image and video tasks, attributing the incremental gain to each optimization. Qwen2-VL is served via vLLM~\cite{kwon2023vllm} with W4A16 quantization; NVILA is deployed with our specialized inference engine. NVILA achieves 1.6--2.2$\times$ faster prefilling and up to 2.8$\times$ higher decoding throughput. All measurements are taken on a single NVIDIA RTX 4090 GPU.}
    \label{fig:inference_efficiency}
\end{figure*}

NVILA achieves competitive accuracy on image and video benchmarks while remaining efficient through the ``\textit{scale-then-compress}'' design. Architecturally, we scale up to native resolution (using 1--12 tiles per image) and then compress visual tokens with a 3$\times$3 spatial-to-channel reshape, yielding a 2.4$\times$ speedup with little accuracy loss. On the data side, we curate a diverse 10M-sample SFT dataset, prune it to a 5M high-quality subset with DeltaLoss, and still consistently outperform LLaVA-OneVision, which trains on more than 8M samples. We further integrate FP8 training for higher throughput, jointly tune ViT/LLM learning rates for fine-tuning efficiency, and apply W8A8 quantization to the vision tower to reduce latency. Together, these full-stack optimizations let NVILA train faster, fine-tune with less memory, and infer with lower latency, all while improving accuracy.

We compare NVILA's inference performance against Qwen2-VL~\cite{wang2024qwen2vl} in \fig{fig:inference_efficiency}. For a fair comparison, both models sample 64 frames per video, with all experiments conducted on a single NVIDIA RTX 4090 GPU. Qwen2-VL is quantized to W4A16 and served with vLLM~\cite{kwon2023vllm}, a state-of-the-art LLM/VLM serving engine. For NVILA, we quantize the LLM backbone to W4A16 and the vision tower to W8A8. With our specialized inference engine, NVILA achieves up to a 2.2$\times$ speedup in prefilling and up to 2.8$\times$ higher decoding throughput over Qwen2-VL.

\section{More Capabilities}

\subsection{Temporal Localization}

Following LITA~\cite{huang2024lita}, we add support for temporal localization in NVILA. We introduce discrete time tokens to represent video timestamps and train the model with a smoothed cross-entropy loss. As shown in \tab{tab:results:lita}, NVILA substantially outperforms all baselines across all metrics.

\begin{table}[t]
    \footnotesize\centering
    \renewcommand{\arraystretch}{1.1}
    \caption{\textbf{Temporal localization}. LITA results are from the original paper; VILA-1.5 results are from our reproduction. NVILA uses the same data mixture as VILA-1.5, with only the base VLM replaced.}
    \begin{tabular}{lrccc}
        \toprule
        & & & \multicolumn{2}{c}{ActivityNet-RTL} \\
        \cmidrule(lr){4-5}
        & & \#Frames & Mean IoU & Precision@0.5 \\
        \midrule
        LITA~\cite{huang2024lita} & 7B & 100 & 24.1 & 21.1 \\
        LITA~\cite{huang2024lita} & 13B & 100 & 28.6 & 25.9 \\
        VILA-1.5 & 8B & 256 & 32.1 & 29.3 \\
        \textbf{NVILA} & 8B & 256 & \textbf{34.8} & \textbf{32.1} \\
        \bottomrule
    \end{tabular}
    \label{tab:results:lita}
\end{table}

\begin{figure}[t]
  \centering
  \includegraphics[width=0.5\textwidth]{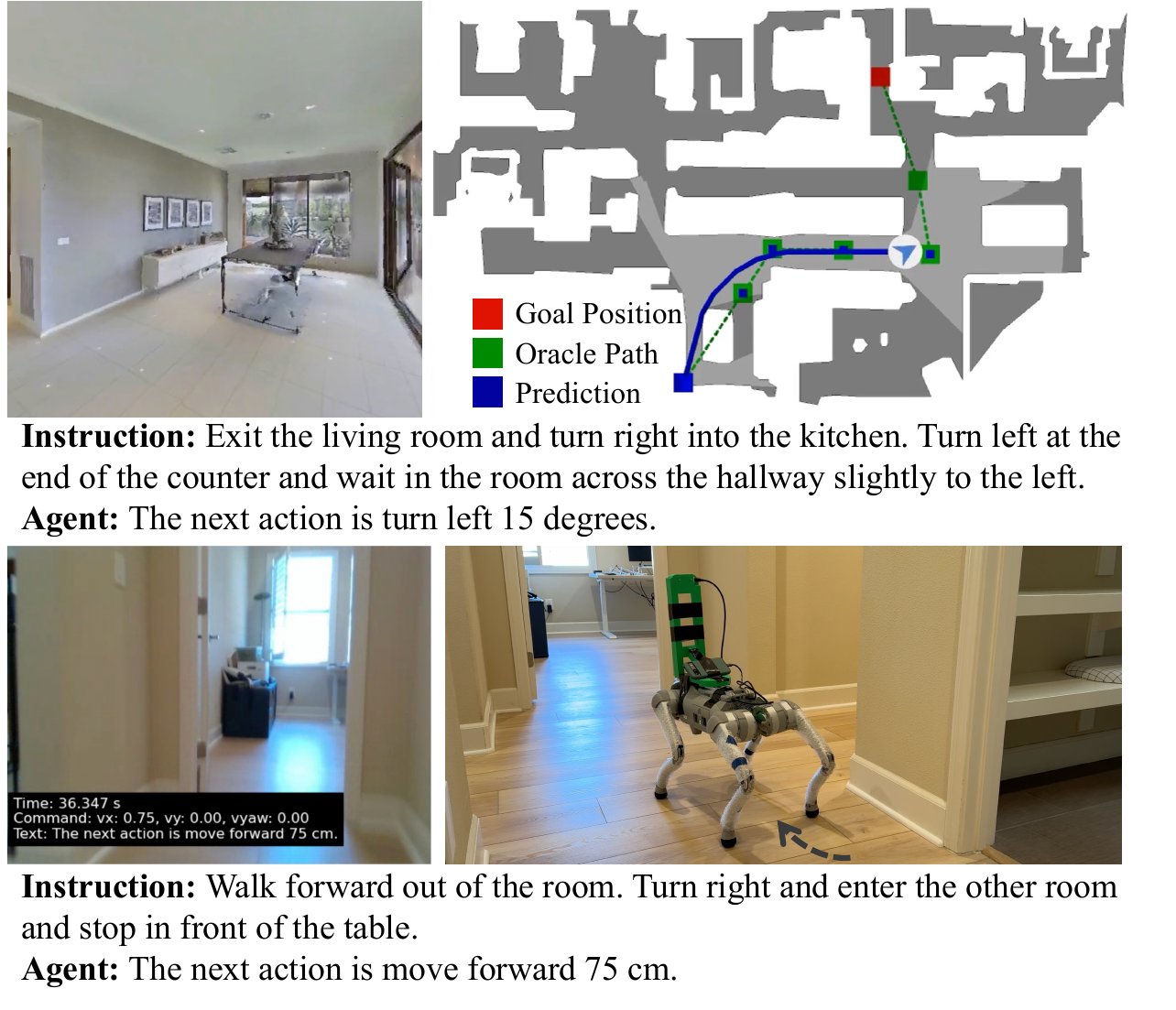}
  \caption{\textbf{Robotic navigation}. NVILA serves as a Vision-Language Navigation agent that takes language instructions and visual observations as input (top: simulation; bottom: real-world). The real-world platform is a Unitree Go2 robot equipped with a LiDAR sensor and an Intel RealSense camera. An RTX 4090 GPU runs NVILA-8B, processing 8 frames per action step.}
  \label{fig:navila}
\end{figure}

\subsection{Robotic Navigation}

NVILA can serve as a strong foundation for robotic agents in Vision-Language Navigation~\cite{krantz2020beyond} and supports real-time deployment on resource-constrained edge devices. At each time step $t$, the agent receives a language instruction and a video observation, plans the next action, and transitions to the next state $t{+}1$, where it receives a new observation. NVILA's efficient and flexible handling of multi-frame inputs enables seamless integration of historical and current observations. The NaVILA framework~\cite{cheng2024navila} introduces a tailored navigation prompt and fine-tunes NVILA using navigation-specific SFT data curated from a simulator~\cite{anderson2018vision}. As shown in \tab{tab:action:navila}, NVILA's straightforward design achieves competitive results on VLN-CE. Real-world deployment results are shown in \fig{fig:navila}: the full camera$\rightarrow$GPU$\rightarrow$action pipeline runs at 1\,Hz.

\begin{table}[t]
    \footnotesize\centering
    \renewcommand{\arraystretch}{1.1}
    \caption{\textbf{Robotic navigation}. All results except NVILA's are from NaVILA~\cite{cheng2024navila}. All models use only RGB inputs. NE: Navigation Error ($\downarrow$); OS: Oracle Success; SR: Success Rate; SPL: Success Rate weighted by Path Length.}
    \begin{tabular}{lrccccc}
        \toprule
        & & & \multicolumn{4}{c}{R2R Val-Unseen} \\
        \cmidrule(lr){4-7}
        & & Obs. & NE $\downarrow$ & OS $\uparrow$ & SR $\uparrow$ & SPL $\uparrow$ \\
        \midrule
        Seq2Seq & -- & RGB & 10.10 & 8.0 & 0.0 & 0.0 \\
        CMA & -- & RGB & 9.55 & 10.0 & 5.0 & 4.0 \\
        NaVid & 7B & RGB & 5.47 & 49.0 & 37.0 & 35.0 \\
        \textbf{NVILA} & 8B & RGB & \bf5.43 & \bf60.4 & \bf53.3 & \bf48.8 \\
        \bottomrule
    \end{tabular}
    \label{tab:action:navila}
\end{table}

\subsection{Medical Application}

NVILA also offers strong potential in the medical domain. The NVILA-M3 framework~\cite{nath2024vilam3} integrates multiple domain-expert models tailored to specific medical tasks, such as image segmentation and classification. These expert models extract and interpret intricate features that are difficult for general VLMs to discern. By coupling specialized models with a vision-language learning paradigm, NVILA-M3 learns nuanced relationships between visual inputs and textual annotations, improving both task-specific accuracy and robustness. As reported in \tab{tab:vilam3}, NVILA-M3 achieves an overall 9\% improvement over task-specific state-of-the-art baselines through the use of expert models. This demonstrates the value of combining domain expertise with general-purpose VLMs in precision-critical fields.

\begin{table}[t!]
    \centering\footnotesize
    \caption{\textbf{Medical application}. Results of NVILA-M3 on medical benchmarks; task-specific baselines and datasets are described in~\cite{nath2024vilam3}. Metrics: accuracy for VQA; BLEU-4 and ROUGE for report generation; F1 for classification.}
    \setlength{\tabcolsep}{4pt}
    \renewcommand{\arraystretch}{1.1}
    \begin{tabular}{lrccccc}
        \toprule
        & & \multicolumn{2}{c}{VQA} & \multicolumn{2}{c}{Report Gen.} & \multicolumn{1}{c}{Classif.} \\
        \cmidrule(lr){3-4}\cmidrule(lr){5-6}\cmidrule(lr){7-7}
        & & Rad & Path & \multicolumn{2}{c}{CXR} & CheXpert \\
        \midrule
        Med-Gemini & -- & 78.8 & 83.3 & 20.5 & 28.3 & 48.3 \\ 
        VILA-M3 & 8B & 84.7 & 91.0 & 21.1 & 32.0 & \textbf{61.6} \\ 
        \textbf{NVILA} & 8B & \textbf{85.5} & \textbf{92.9} & \textbf{22.8} & \textbf{32.8} & 61.1 \\ 
        \midrule
        \multicolumn{2}{l}{\textcolor{black!50}{Task-spfc. SOTA}} & \textcolor{black!50}{84.2} & \textcolor{black!50}{91.7} & \textcolor{black!50}{15.4} & \textcolor{black!50}{30.6} & \textcolor{black!50}{51.5} \\ 
        \bottomrule
    \end{tabular}
    \label{tab:vilam3}
\end{table}
\section{Related Work}

\subsection{Visual Language Models}

VLMs, especially proprietary ones, have advanced rapidly over the past two years. OpenAI upgraded from GPT-4V~\cite{openai2023gpt4v} to GPT-4o~\cite{openai2024gpt4o}, achieving 5--10\% gains across image and video QA benchmarks. Google extended the context length to 1M tokens in Gemini~1.5~Pro~\cite{google2024gemini}, a significant step up from Gemini~1.0~\cite{google2023gemini}; Gemini~1.5~Pro tops the Video-MME leaderboard~\cite{fu2024videomme} for long-video understanding. Anthropic released Claude~3.5~\cite{anthropic2024claude35}, which is competitive with GPT-4o and improves notably over Claude~3~\cite{anthropic2024claude3}. Other proprietary models have followed similar trajectories, including Apple's MM1~$\to$~MM1.5~\cite{zhang2024mm15} and xAI's Grok-1.5~\cite{xai2024grok15}~$\to$~Grok-2~\cite{xai2024grok2}.

Meanwhile, open-source VLMs continue to evolve at both the system/framework level~\cite{kuchaiev2019nemo} and the algorithm/recipe level~\cite{lin2024vila}, progressively narrowing the gap to proprietary models~\cite{xue2024longvila,fang2024vila2,shi2024eagle,dai2024nvlm,wang2024qwen2vl}. Many recent open VLMs now match or exceed GPT-4V and GPT-4o on key benchmarks. Representative examples include InternVL2~\cite{chen2024internvl}, Qwen2-VL~\cite{wang2024qwen2vl}, LLaVA-OneVision~\cite{li2024llavaonevision}, Llama~3.2~Vision~\cite{meta2024llama3}, Molmo~\cite{deitke2024molmo}, NVLM~\cite{dai2024nvlm}, and MiniCPM-V~\cite{yao2024minicpmv}.

Despite this progress in accuracy, much less attention has been paid to the efficiency of training, fine-tuning, and deployment. This paper explores how to build VLMs that are both highly accurate and optimized for end-to-end efficiency.

\subsection{Efficiency}

\xpar{Token reduction.}
Prior work~\cite{bolya2023tome,chen2024fastv,xiong2024pyra,shen2024longvu,choi2024vidtldr,jin2024chatunivi,xu2024slowfast,cheng2024spatialrgpt} has studied spatial and temporal token reduction, but none focuses on reducing tokens for a frontier VLM, where preserving accuracy at scale is the primary challenge.

\xpar{Data selection.}
For LLM pre-training, promising approaches include domain mixing~\cite{xie2023doremi}, sample-wise data selection~\cite{xia2024less,du2023mods}, and theory-driven optimal selection~\cite{gu2024pds}. Comparatively few studies tackle the mixed image-text setting of VLM training; in this work we specifically target supervised fine-tuning (SFT) data for VLMs.

\xpar{Low-precision training.}
FP8 training~\cite{fishman2024scaling,micikevicius2022fp8} has gained traction for LLMs, but to our knowledge no prior work has demonstrated its feasibility for VLMs without sacrificing accuracy.

\xpar{Pruning, distillation, and quantization.}
Pruning and distillation~\cite{muralidharan2024minitron,dery2024bonsai} are commonly applied to LLMs, but their best practice for VLMs remains open: should the LLM be pruned or distilled before adding a vision encoder, or should the assembled VLM be pruned or distilled? Similarly, quantization methods such as AWQ~\cite{lin2024awq} and GPTQ~\cite{frantar2022gptq} are well-studied for LLMs, and VILA~\cite{lin2024vila} has shown that AWQ transfers directly to VLMs. However, little attention has been paid to quantizing the vision encoder, which becomes critical at higher input resolutions and longer videos.

\xpar{Parameter-efficient fine-tuning.}
Methods such as LoRA~\cite{hu2021lora}, DoRA~\cite{liu2024dora}, QLoRA~\cite{dettmers2024qlora}, and GaLore~\cite{zhao2024galore} are widely used to reduce LLM fine-tuning memory. For VLMs, which combine a vision encoder with an LLM, efficient fine-tuning techniques are still underexplored, a gap we address in this work.
\section{Conclusion}

This paper introduced NVILA, a family of open VLMs designed to strike a strong balance between efficiency and accuracy. By adopting the ``\textit{scale-then-compress}'' paradigm, NVILA processes high-resolution images and long videos efficiently while maintaining high accuracy. We further systematically optimize its efficiency across the entire lifecycle, from training and fine-tuning to deployment. NVILA matches or surpasses current leading VLMs in accuracy while being substantially more resource-efficient, and it opens up new possibilities for applications such as temporal localization, robotic navigation, and medical imaging. We hope NVILA will empower researchers and developers across a wide range of applications and research directions.

{
\small
\bibliographystyle{unsrt}
\bibliography{main}
}

\appendix
\renewcommand{\thesection}{A.\arabic{section}}
\renewcommand{\thetable}{A\arabic{table}}
\renewcommand{\thefigure}{A\arabic{figure}}
\setcounter{section}{0}
\setcounter{table}{0}
\setcounter{figure}{0}

\clearpage
\begin{table*}[t]
    \footnotesize\centering
    \renewcommand{\arraystretch}{1.1}
    \caption{\textbf{Data recipe}. Training data for each stage, grouped by category.}
    \begin{tabular}{rp{360pt}}
        \toprule
        \rowcolor{black!10} \multicolumn{2}{l}{\textit{\textbf{Stage 1: Projector Initialization}}} \\
        \midrule
        Feature Align & LLaVA-CC3M-Pretrain~\cite{liu2024llava} \\
        \midrule\midrule
        \rowcolor{black!10} \multicolumn{2}{l}{\textit{\textbf{Stage 2: Visual Encoder Pre-Training}}} \\
        \midrule
        Recaptioned Data & ALLAVA~\cite{chen2024allava}  \\ \midrule
        Document & Docmatix~\cite{laurencon2024idefics3}, PDFA~\cite{montalvo2024pdfa} \\ 
        \midrule
        OCR & LSVT~\cite{sun2019lsvt}, ArT~\cite{chng2019art} \\
        \midrule\midrule
        \rowcolor{black!10} \multicolumn{2}{l}{\textit{\textbf{Stage 3: Token Processor Pre-Training}}} \\
        \midrule
        Recaptioned Data & {COYO~\cite{byeon2022coyo} (25M Subset and recaptioned by VILA$^2$~\cite{fang2024vila2}), ShareGPT4v-Pretrain~\cite{chen2023sharegpt4v}}  \\ \midrule
        Document & Docmatix~\cite{laurencon2024idefics3}, UniChart-Pretrain~\cite{masry2023unichart} \\ \midrule
        Interleaved Data & MMC4~\cite{zhu2024multimodal} \\
        \midrule\midrule
        \rowcolor{black!10} \multicolumn{2}{l}{\textit{\textbf{Stage 4: Image Instruction-Tuning}}} \\
        \midrule
        Hybrid     & ShareGPT4V-SFT~\cite{chen2023sharegpt4v}, Molmo(subset)~\cite{deitke2024molmo}, The Cauldron(subset)~\cite{laurencon2024idefics2}, Cambrian(subset)~\cite{tong2024cambrian}, LLaVA-OneVision(subset)~\cite{li2024llavaonevision}   \\
        \midrule
        Captioning & {MSR-VTT~\cite{xu2016msrvtt},  Image Paragraph Captioning~\cite{krause2017hierarchical}, ShareGPT4V-100K~\cite{chen2023sharegpt4v}} \\
        \midrule
        Reasoning  & CLEVR~\cite{johnson2017clevr}, NLVR2~\cite{suhr2019nlvr2}, VisualMRC~\cite{tanaka2021visualmrc} \\
        \midrule
        Document &  DocVQA~\cite{mathew2021docvqa}, UniChart-SFT~\cite{masry2023unichart}, ChartQA~\cite{masry2022chartqa} \\ \midrule
        OCR & TextCaps~\cite{sidorov2020textcaps}, OCRVQA~\cite{mishra2019ocr}, ST-VQA~\cite{biten2019scene}, POIE~\cite{kuang2023visual}, SROIE~\cite{huang2019icdar2019}, SynthDoG-en~\cite{kim2022donut},  TextOCR-GPT4V, ArxivQA~\cite{li2024multimodal}, LLaVAR~\cite{zhang2023llavar} \\
        \midrule
        General VQA & ScienceQA~\cite{lu2022learn}, VQAv2~\cite{antol2015vqa}, ViQuAE~\cite{lerner2022viquae}, Visual Dialog~\cite{das2017visual}, GQA~\cite{hudson2019gqa}, Geo170K~\cite{gao2023gllava}, LRV-Instruction~\cite{liu2024mitigating}, RefCOCO~\cite{yu2016modeling}, GeoQA~\cite{chen2021geoqa}, OK-VQA~\cite{marino2019okvqa}, TabMWP~\cite{lu2023promptpg}, EstVQA~\cite{wang2020general} \\
        \midrule
        Diagram \& Dialogue & DVQA~\cite{kafle2018dvqa}, AI2D~\cite{kembhavi2016diagram}, Shikra~\cite{chen2023shikra}, UniMM-Chat~\cite{yu2023unimmchat} \\
        \midrule
        Instruction & LRV-Instruction~\cite{liu2024mitigating}, SVIT~\cite{zhao2023svit}, MMC-Instruction~\cite{liu2024mmc},
        MM-Instruction~\cite{liu2024mminstruct} \\
        \midrule
        Text-only & FLAN-1M~\cite{wei2022finetuned}, MathInstruct~\cite{yue2024mammoth}, Dolly~\cite{databricks2023dolly}, GSM8K-ScRel-SFT~\cite{yuan2023scaling} \\
        \midrule
        Knowledge & WordART~\cite{xie2022toward}, WIT~\cite{srinivasan2021wit}, STEM-QA~\cite{shen2024measuring} \\
        \midrule
        Medical & PathVQA~\cite{he2020pathvqa}, SLAKE~\cite{liu2021slake}, MedVQA \\
\midrule
        Video & ActivityNet-QA~\cite{yu2019activitynetqa}, MSRVTT-QA~\cite{xu2016msrvtt}, iVQA~\cite{yang2021just}, Youcook2~\cite{zhou2018towards}, VaTeX~\cite{wang2019vatex}, ShareGPTVideo~\cite{zhang2024direct}  \\
        \midrule
        \rowcolor{black!10} \multicolumn{2}{l}{\textit{\textbf{Stage 5: Video Instruction-Tuning}}} \\
        \midrule
        Video & LLaVA-Video-178K~\cite{zhang2024llavavideo} \\
        \midrule
        Image & LLaVA-OneVision(subset)~\cite{li2024llavaonevision} \\
        \bottomrule
    \end{tabular}
    \label{tab:recipe:data}
\end{table*}

\end{document}